\documentclass{llncs}
\usepackage{makeidx}  
\usepackage{verbatim}
\usepackage{paralist}
\usepackage[pdftex]{graphicx}
\usepackage{subcaption}

\usepackage[pdftex]{hyperref}
\hypersetup{
    unicode=false,          
    pdffitwindow=false,     
    pdfstartview={FitH},    
    pdftitle={A CNL for C-O Diagrams},    
    pdfauthor={Camilleri, Paganelli, Schneider},     
    colorlinks=true,       
    linkcolor=black,          
    citecolor=black,        
    filecolor=black,      
    urlcolor=black           
}

\usepackage{xcolor}

\usepackage{tikz}
\usetikzlibrary{arrows,shapes, shapes.geometric, shapes.symbols, shapes.arrows, shapes.multipart, shapes.callouts, shapes.misc}

\usepackage{amsmath}
\usepackage{textcomp}

\definecolor{lstbg}{gray}{0.95}
\usepackage{listings}
\lstdefinestyle{cnl-case-study}{
  captionpos=b,                   
  morekeywords={},
}

\newcommand{\codiag}{\textit{C-O Diagram}}
\newcommand{\codiags}{\textit{C-O Diagrams}}
\newcommand{\COML}{COML}
\newcommand{\coml}{COML}
\newcommand\CL{\ensuremath{\mathcal{CL}}}
\newcommand\anacon{\ensuremath{\mathsf{AnaCon}}}

\newcommand{\conbox}[1]{{\tt #1}}
\newcommand{\convar}[1]{{\tt #1}}
\newcommand{\conname}[1]{{\tt #1}}
\newcommand{\COdiaOp}[1]{{\tt #1}}

\begin{document}

\pagestyle{headings}  

\mainmatter              

\title{A CNL for Contract-Oriented Diagrams}

\titlerunning{A CNL for \codiags}

\author{John J. Camilleri \and Gabriele Paganelli \and Gerardo Schneider}

\authorrunning{Camilleri, Paganelli and Schneider} 

\institute{Department of Computer Science and Engineering, \\
  Chalmers University of Technology and the
  University of Gothenburg, Sweden \\
  \email{\{john.j.camilleri@cse.gu.se, gabpag@chalmers.se, gerardo@cse.gu.se\}}}

\maketitle

\begin{abstract}
We present a first step towards a framework for defining and manipulating normative documents or contracts described as \textit{Contract-Oriented (C-O) Diagrams}.
These diagrams provide a visual representation for such texts, giving the possibility to express a signatory's obligations, permissions and prohibitions, with or without timing constraints, as well as the penalties resulting from the non-fulfilment of a contract.
This work presents a CNL for verbalising \codiags, a web-based tool allowing editing in this CNL, and another for visualising and manipulating the diagrams interactively.
We then show how these proof-of-concept tools can be used by applying them to a small example.

\keywords{normative texts, electronic contracts, c-o diagrams, controlled natural language, grammatical framework}
\end{abstract}


\section{Introduction and background}

Formally modelling normative texts such as legal contracts and regulations is not new.
But the separation between logical representations and the original natural language texts is still great.
CNLs can be particularly useful for specific domains where the coverage of full language is not needed, or at least when it is possible to abstract away from some irrelevant aspects.

In this work we take the \codiag\ formalism for normative documents \cite{DCM+13svn},
which specifies a visual representation and logical syntax for the formalism, together with a translation into timed automata.
This allows model checking to be performed on the modelled contracts.
Our concern here is how to ease the process of writing and working with such models,
which we do by defining a CNL which can translate unambiguously into a \codiag.
Concretely, the contributions of our paper are the following:
\begin{compactenum}
\item Syntactical extensions to \codiags~concerning executed actions and cross-references (section \ref{sec:extension});
\item A CNL for \codiags\ implemented using the Grammatical Framework (GF), precisely mapping to the formal grammar of the diagrams (section \ref{sec:cnl}).
\item Tools for visualising and manipulating \codiags\ (section \ref{sec:architecture}):
\begin{compactenum}
\item A web-based visual editor for \codiags;
\item A web-based CNL editor with real-time validation;
\item An XML format \COML\ used as a storage and interchange format.
\end{compactenum}
\end{compactenum}
We also present a small example to show our CNL in practice (section \ref{sec:case-study})
and an an initial evaluation of the CNL (section \ref{sec:evaluation}).
In what follows we provide some background for \codiags\ and GF.



\subsection{\codiags}
Introduced by
Mart\'{i}nez et al. \cite{MCD+10mvs},
\codiags\ provide a means for visualising normative texts containing the modalities of obligation, permission and prohibition.
They allow the representation of complex clauses describing these norms for different signatories, as well as {\it reparations} describing what happens when obligations and prohibitions are not fulfilled.

The basic element is the {\it box} (see \autoref{fig:examples}), representing a basic contract clause.
A box has four components:
\begin{inparaenum}[i)]
\item \textit{guards} specify the conditions for enacting the clause;
\item \textit{time restrictions} restrict the time frame during which the contract clause must be satisfied;
\item the \textit{propositional content} of a box specifies a modality applied over actions, and/or the actions themselves;
\item a \textit{reparation}, if specified, is a reference to another contract that must be satisfied in case the main norm is not.
\end{inparaenum}
Each box also has an \textit{agent} indicating the performer of the action,
and a unique \textit{name} used for referencing purposes.
Boxes can be expanded by using three kinds of refinement: {\it conjunction}, {\it choice}, and {\it sequencing}.

\begin{figure}[t]
  \centering
  \begin{align*}
    C :=\;&(agent, name, g, tr, O(C_2), R) \\
       |\;&(agent, name, g, tr, P(C_2), \epsilon) \\
       |\;&(agent, name, g, tr, F(C_2), R) \\
       |\;&(\epsilon, name, g, tr, C_1, \epsilon) \\
    C_1 :=\;&C\;(And\;C)^+\;|\;C\;(Or\;C)^+\;|\;C\;(Seq\;C)^+\;|\;Rep(C) \\
    C_2 :=\;&a\;|\;C_3\;(And\;C_3)^+\;|\;C_3\;(Or\;C_3)^+\;|\;C_3\;(Seq\;C_3)^+ \\
    C_3 :=\;&(\epsilon, name, \epsilon, \epsilon, C_2, \epsilon) \\
    R :=\;&C\;|\;\epsilon
  \end{align*}
  \caption{Formal syntax of \codiags~\cite{DCM+13svn}} 
  \label{eqn:codiag-syntax}
\end{figure}

The diagrams have a formal definition given by the syntax shown in \autoref{eqn:codiag-syntax}.
For an example of a \codiag, see \autoref{fig:final-codiag} (this example will be explained in more detail in \autoref{sec:case-study}).


\subsection{Grammatical Framework}

GF \cite{GF} is both a language for multilingual grammar development and a type-theoretical logical framework,
which provides a mechanism for mapping abstract logical expressions to a concrete language.
With GF, the language-independent structure of a domain can be encoded in the abstract syntax, while language-specific features can be defined in potentially multiple concrete languages.
Since GF provides both a {\it parser} and {\it lineariser} between concrete and abstract languages, multi-lingual translation can be achieved using the abstract syntax as an interlingua.

GF also comes with a standard library called the {\it Resource Grammar Library} (RGL) \cite{gf-rgl}.
Sharing a common abstract syntax, this library contains implementations of over 30 natural languages.
Each resource grammar deals with low-level language-specific details such as word order and agreement.
The general linguistic descriptions in the RGL can be accessed by using a common language-independent API.
This work uses the English resource grammar, simplifying development and making it easier to port the system to other languages.


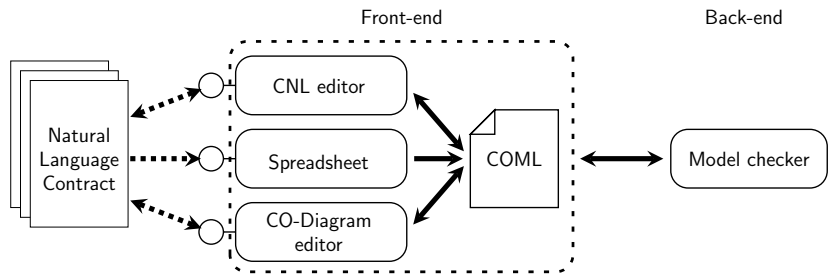
\begin{figure}[t]
  \centering
    \scalebox{0.65}{
      \newcommand{\backendhook}{6.5}
\newcommand{\systemstart}{-0.4}
\newcommand{\contracthook}{2.3}
\newcommand{\systembottom}{-3}
\newcommand{\systemtop}{0}

\sffamily\large
\begin{tikzpicture}[>=stealth]
	\def\xmlfile{node{COML}
	  ++(-0.9,0)
	  node (centerxml){}
	  node (upxml){}
	  node (downxml){} --
	  ++(0,0.5) --
	  ++(0.5,0.5) --
	  ++(0,-0.5) --
	  ++(-0.5,0) --
	  ++(0.5,0.5) --
	  ++(1.3,0) --
	  ++(0,-2) --
	  ++(-1.8,0) -- cycle
	}
	\newcommand{\FEcomponent}[2]{++(0,0) rectangle ++(3.5,1.2) ++(-3.5,-0.6) -- ++(-0.25,0) ++(-0.25,0) circle (0.25) node[shape=circle](#2){} ++(2.2,0)node[align=center,rectangle]{#1} ++(1.8,0) node(#2inner){}}
	\newcommand{\Modelchecker}{+(0,0.6)node(backendint){} ++(0,0) rectangle ++(3.2,1.2) ++(-1.6,-0.6)node[align=center,rectangle]{Model checker};}
	\newcommand{\NLcontract}[2]{
	\foreach \x in {(#1,#2), (#1+0.2,#2-0.2), (#1+0.4,#2-0.4)}
	  \draw[fill=white] \x rectangle ++(2,3);

	\draw node[align=center, rectangle](nlc) at (#1+1.4,#2+1){Natural\\Language\\Contract};
	\draw node[align=center](nlcup) at (#1+\contracthook,#2+0.2) {};
	\draw node[align=center](nlccenter) at (#1+\contracthook,#2+1) {};
	\draw node[align=center](nlcdown) at (#1+\contracthook,#2+1.8) {};
	}

	\draw[color=white] (\systemstart,\systembottom) rectangle (12.5,\systemtop);
	\draw[loosely dashed, ultra thick, rounded corners= 10pt] (\systemstart,-1.6,0.3) rectangle (\backendhook,3);
	\draw[thick, fill=white, rounded corners= 10pt] (\systemstart,1.5) \FEcomponent{CNL editor}{cnl};
	\draw[thick, fill=white, rounded corners= 10pt] (\systemstart,0) \FEcomponent{\\Spreadsheet}{google};
	\draw[thick, fill=white, rounded corners= 10pt] (\systemstart,-1.5) \FEcomponent{CO-Diagram\\editor}{diagram};
	\draw[thick, fill=white] (5.3,0.6) \xmlfile;
	\draw[thick, fill=white, rounded corners= 10pt] (8.5,0) \Modelchecker;
	\node[align=center] at (3.0,3.5) {Front-end};
	\NLcontract{-5}{-0.4};
	\node[align=center] at (10,3.5) {Back-end};
	\draw[<->,line width=3, dashed] (nlcup) -- (diagram);
	\draw[->,line width=3, dashed] (nlccenter) -- (google);
	\draw[<->,line width=3, dashed] (nlcdown) -- (cnl);
	\draw[<->,line width=3] (cnlinner) -- (upxml);
	\draw[->,line width=3] (googleinner) -- (centerxml);
	\draw[<->,line width=3] (diagraminner) -- (downxml);
	\node(backendhook) at (\backendhook,0.6) {};
	\draw[<->, line width=3](backendhook) -- (backendint);

\end{tikzpicture}
    }
    \vspace{-5mm}
    \caption{The contract processing framework. Dashed arrows represent manual interaction, solid ones automated interaction.}
    \label{fig:architecture}
\end{figure}

\section{Implementation}\label{sec:architecture}

\subsection{Architecture}
The contract processing framework presented in this work is depicted in \autoref{fig:architecture}.
There is a {\em front-end} concerned with the modelling of contracts in a formal representation,
and a {\em back-end} which uses formal methods to detect conflicts, verify properties, and process queries about the modelled contract.
The back-end of our system is still under development, and involves the automatic translation of contracts into timed automata which can be processed using the UPPAAL tool \cite{Larsen2014}.

The front-end, which is the focus of this paper, is a collection of web tools that communicate using our XML format named \COML.%
\footnote{An example of the format, together with an XSD schema defining the structure, is available online at \url{http://remu.grammaticalframework.org/contracts}}
This format closely resembles the \codiag\ syntax (\autoref{eqn:codiag-syntax}).
The tools in our system allow a contract to be expressed as a CNL text, spreadsheet, and \codiag.
Any modification in the diagram is automatically verbalised in CNL and vice versa.
A properly formatted spreadsheet may be converted to a \COML\ file readable by the other editors.
These tools use HTML5 \cite{Navara:14:H} local storage for exchanging data.

\subsubsection{Translation process}

\begin{figure}[t]
  \centering
  \includegraphics[width=\textwidth]{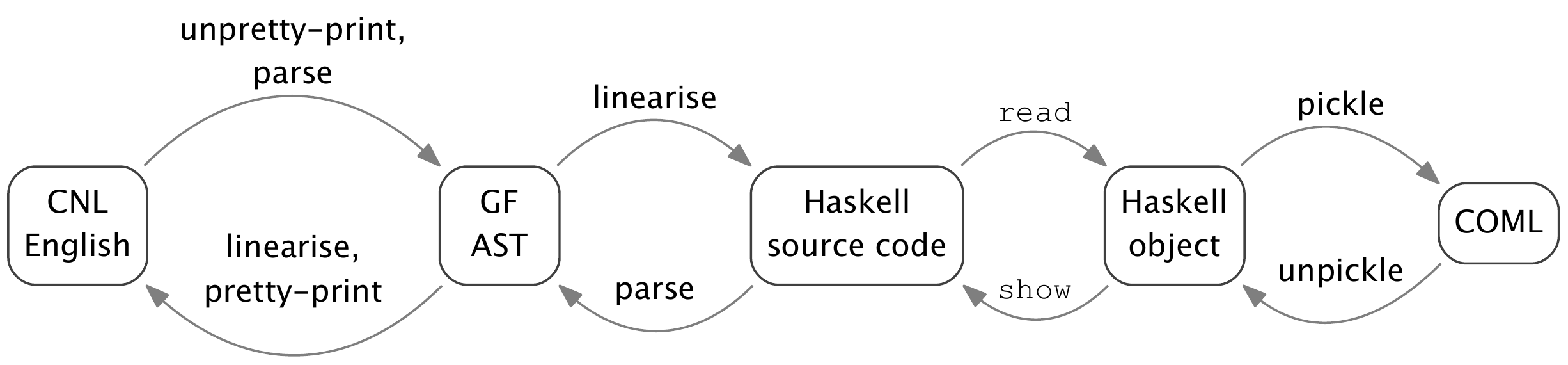}
  \caption{Conversion process from CNL to \coml\ and back.}
  \label{fig:conversion}
\end{figure}

The host language for all our tools is Haskell,
which allows us to define a central data type precisely reflecting the formal \codiag\ grammar (\autoref{eqn:codiag-syntax}).
We also define an abstract syntax in GF which closely matches this data type, and translate between CNL and Haskell source code via two concrete syntaxes.
As an additional processing step after linearisation with GF,
the generated output is passed through a pretty-printer, adding newlines and indentations as necessary (\autoref{sec:cnl-lists}).
%
The Haskell source code generated by GF can be converted to and from actual objects by deriving the standard \verb|Show| and \verb|Read| type classes.
Conversion to the \COML\ format is then handled by the HXT library, which generates both a parser and generator from a single {\em pickler} function.
The entire process is summarised in \autoref{fig:conversion}.


\subsection{Editing tools}

The visual editor allows users to visually construct and edit \codiags\ of the type seen in \autoref{sec:case-study}.
It makes use of the mxGraph JavaScript library 
providing the components of the visual language and several facilities such as converting and sending the diagram to the CNL editor, validation of the diagram, conversion to PDF and PNG format.

The editor for CNL texts uses the ACE JavaScript library 
to provide a text-editing interface within the browser.
The user can verify that their CNL input is valid with respect to grammar, by calling the GF web service.
Errors in the CNL are highlighted to the user.
A valid text can then be translated into \COML\ with the push of a button.


\subsection{Syntactic extensions to \codiags}\label{sec:extension}

This work also contributes two extensions to \codiag\ formalism:
\begin{compactenum}
\item To the grammar of guards, we have add a new condition on whether an action $a$ has been performed ($done(a)$);
\item We add also a new kind of box for cross-references.
This enhances \codiags~with the possibility to have a more modular way to ``jump'' to other clauses.
This is useful for instance when referring to {\it reparations}, and to allow more general cases of ``repetition''.
\end{compactenum}
\noindent
Our tool framework also includes some additional features for facilitating the manipulation of \codiags.
The most relevant to the current work is the automatic generation of clocks for each action.
This is done by implicitly creating a clock \verb+t_name+ for each box \verb+name+.
When the action or sub-contract \verb+name+ is completed, the clock \verb+t_name+ is reset,
allowing the user to refer to the time elapsed since the completion of a particular box.


\section{CNL}\label{sec:cnl}

This section describes some of the notable design features of our CNL.
Examples of the CNL can be found in the example in \autoref{sec:case-study}.

\subsection{Grammar}

The GF abstract syntax matches closely the Haskell data type designed for \codiags, with changes only made to accommodate GF's particular limitations.
Optional arguments such as guards are modelled with a category \verb|MaybeGuard| having two constructors \verb|noGuard| and \verb|justGuard|, where the latter is a function taking a list of guards, \verb|[Guard]|.
The same solution applies to timing constraints.
Since GF does not have type polymorphism, it is not possible to have a generalised \verb|Maybe| type as in Haskell.
To avoid ambiguity, lists themselves cannot be empty; the base constructor is for a singleton list.

In addition to this core abstract syntax covering the \codiag\ syntax, the GF grammar also imports phrase-building functions from the RGL,
as well as the large-scale English dictionary \verb|DictEng| containing over 64,000 entries.



\subsection{Language features}

\subsubsection{Contract clauses}

A simple contract verbalisation consists of an \textbf{agent}, \textbf{modality}, and an \textbf{action}, corresponding to the standard subject, verb and object of predication.
The modalities of obligation, permission and prohibition are respectively indicated by the keywords \verb|required|, \verb|may| (or \verb|allowed| when referring to complex actions) and \verb|mustn't| (or \verb|forbidden|).

Agents are noun phrases (NP), while actions are formed from either an intransitive verb (V), or a transitive verb (V2) with an NP representing the object.
This means that every agent and action must be a grammatically-correct NP/VP, built from lexical entries found in the dictionary and phrase-level functions in the RGL.
This allows us to correctly inflect the modal verb according to the agent (subject) of the clause:
\begin{lstlisting}
1 : Mary is required to pay
2 : Mary and John are required to pay
\end{lstlisting}

\subsubsection{Constraints}
The arithmetic in the \codiag\ grammar covering guards and timing restrictions is very general, allowing the usual comparison operators between variable or clock names and values, combined with operators for negation and conjunction.
Their linearisation can be seen in line 9 of \autoref{fig:final-cnl}.




Each contract clause in a \codiag\ has an implicit timer associated with it called \verb|t_name|, which is reset when the contract it refers to is completed.
These can be referred to in any timing restriction,
effectively achieving relative timing constraints by referring to the time elapsed since the completion of another contract.

\subsubsection{Conjunction}
\label{sec:cnl-lists}

Multiple contracts can be combined by conjunction, choice and sequencing.
GF abstract syntax supports lists, but linearising them into CNL requires special attention.
Lists of length greater than two must be bulleted and indented, with the entire block prefixed with a corresponding keyword:
\begin{lstlisting}
1 : all of
  - 1a : Mary may eat a bagel
  - 1b : John is required to pay
\end{lstlisting}
When unpretty-printed prior to parsing, this is converted to:
\begin{lstlisting}
1 : all of { - 1a : Mary ... bagel - 1b : John ... pay }
\end{lstlisting}
For a combination of exactly two contracts, the user has the choice to use the bulleted syntax above, or inline the clauses directly using the appropriate combinator, e.g. \verb|or| for choice.
This applies to combination of contracts, actions and even guards and timing restrictions.

In the case of actions the syntax is slightly different since there is a single modality applied to multiple actions. 
Here, the actions appear in the infinitive form and the combination operator appears at the end of each line (except the final one):
\begin{lstlisting}
2 : Mary is allowed
  - 2a : to pay , or
  - 2b : to eat a bagel
\end{lstlisting}
This list syntax allows for nesting to an arbitrary depth.

\subsubsection{Names}
The \codiag\ grammar dictates that all contract clauses should have a name ({\em label}).
These provide modularity by allowing referencing of other clauses by label, e.g. in reparations and relative timing constraints.
Since the CNL cannot be lossy with respect to the \COML, these labels appear in the CNL linearisation too (see \autoref{fig:final-cnl}).
Clause names are free strings, but must not contain any spaces. This avoids the need for double quotes in the CNL.
These labels do reduce naturalness somewhat, but we believe that this inconvenience can be minimised with the right editing tool.


\section{Coffee machine example}
\label{sec:case-study}

A user Eva must analyse the following description of the operation of a coffee machine, and construct a formal model for it.
She will do this interactively, switching between editing the CNL and the visual representation.

{\footnotesize\em
\begin{quote}
   To order a drink the client inputs money and selects a drink. 
   Coffee can be chosen either with or without milk.
   The machine proceeds to pour the selected drink, provided the money paid covers its price, returning any change.
   The client is notified if more money is needed;
   they may then add more coins or cancel the order.
   If the order is cancelled or nothing happens after 30 seconds, the money is returned.
   The machine only accepts euro coins.
\end{quote}}
%

\begin{figure}[t]
  \begin{subfigure}[b]{0.5\textwidth}
  \centering
    \includegraphics[height=2.7cm]{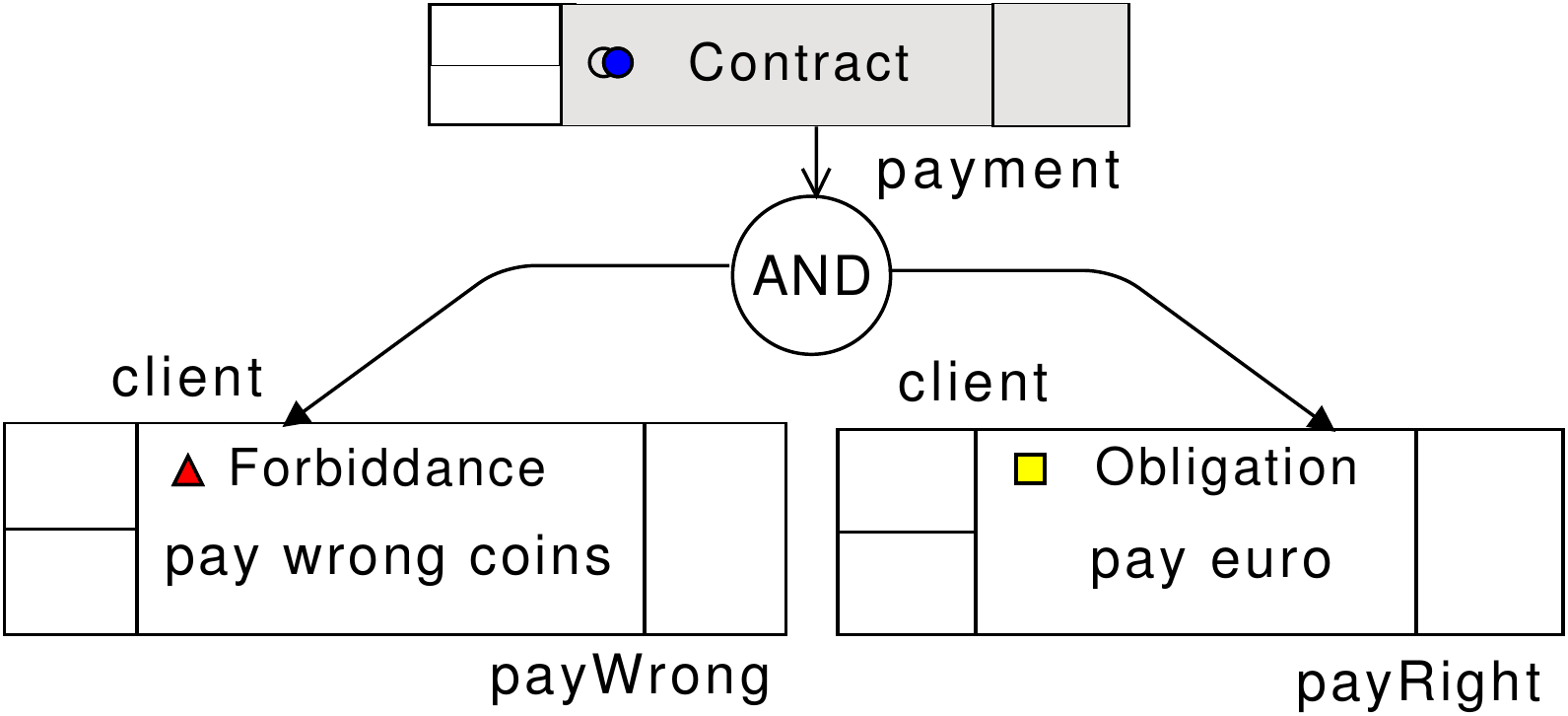}
    \caption{Payment options}
    \label{fig:examples:payment-and}
  \end{subfigure}%
  \begin{subfigure}[b]{0.5\textwidth}
    ~~~~
    \includegraphics[height=3.0cm]{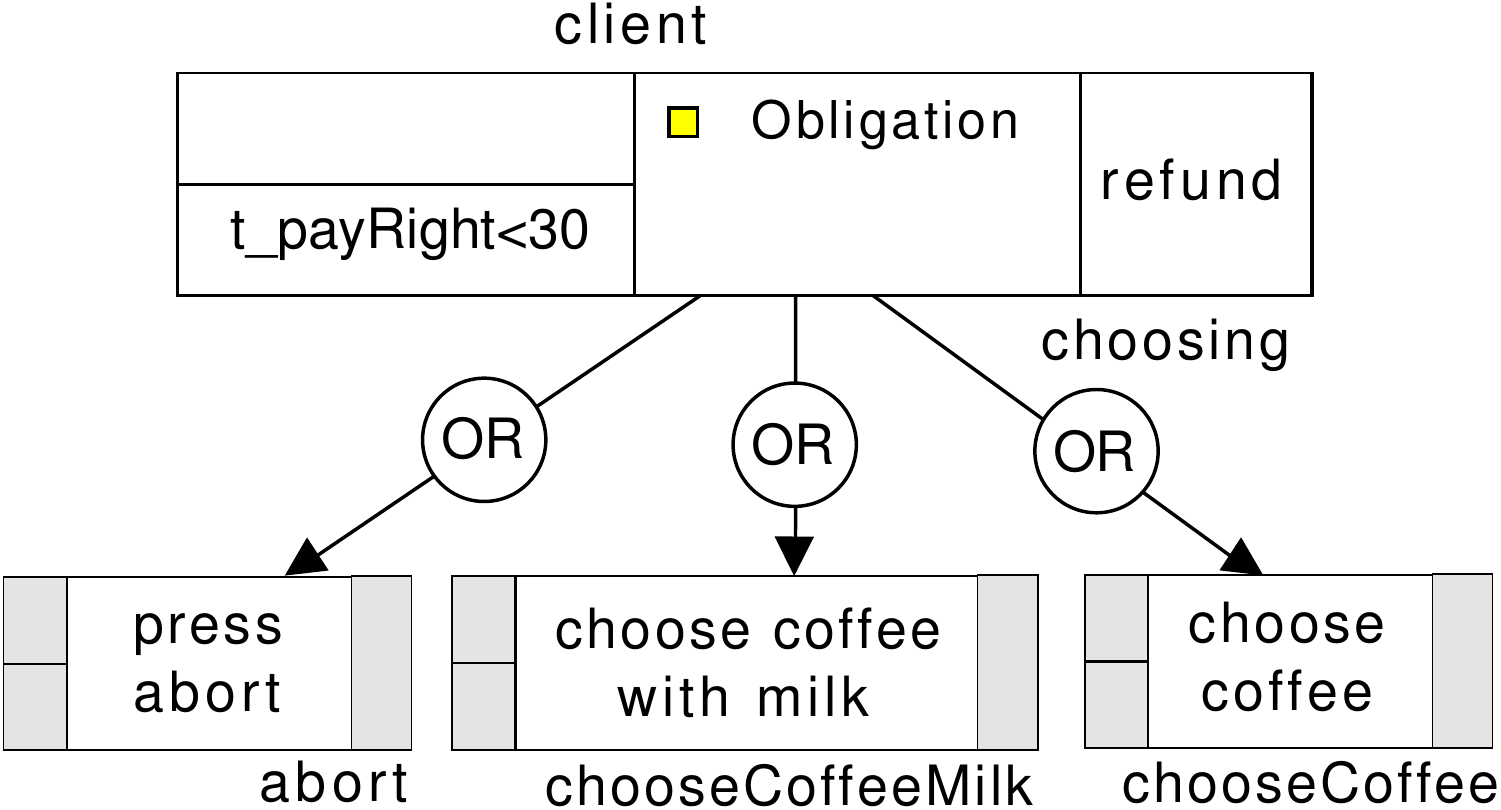}
    \caption{Choices in selection}
    \label{fig:examples:choosing}
  \end{subfigure}
  \begin{subfigure}[b]{\textwidth}
\begin{lstlisting}[style=cnl-case-study,numbers=left]
payment : 
  payWrong : client mustn't pay wrong coins otherwise see refund and 
  payRight : client is required to pay euro
choosing : when clock t_payRight less than 30 client is required
    - abort : to press abort , or
    - chooseCoffeeMilk : to choose coffee with milk , or
    - chooseCoffee : to choose coffee  otherwise see refund
\end{lstlisting}
    \label{fig:cnl:refinements}
  \end{subfigure}
  \caption{Different kinds of complex contracts and their verbalisation.}
  \label{fig:examples}
\end{figure}

Eva first needs to identify:
\begin{inparaenum}[i)]
\item the \emph{actors} (client and machine),
\item the \emph{actions} (pay, accept, select, pour, refund),
\item and the \emph{objects} (beverage, money, timer).
\end{inparaenum}
The first sentence suggests that to obtain a drink the client {\it must} insert coins.
Eva therefore drops an \conbox{obligation} box in the diagram editor and fills the name, agent and action fields.
Only accepting euro is modelled as a prohibition to the client using a \conbox{forbiddance} box.
The two boxes are linked using a \conbox{contract} box as shown in \autoref{fig:examples:payment-and}.

Eva now wants to model the choice of beverage, and the possibility the aborting of the process.
She creates an \conbox{obligation} box named \conname{choosing},
adding the timed constraint \convar{t\_payRight < 30} to model the 30 second timeout.
She then appends two action boxes using the \COdiaOp{or} refinement, corresponding to the choice of drinks (see \autoref{fig:examples:choosing}).
Eva translates the diagram to CNL and modifies the text, adding the action \conname{abort : to press abort} as a refinement of \conname{choosing}.
The result is shown in line 4 of \autoref{fig:final-cnl}.



\begin{figure}[p]
  \includegraphics[width=\textwidth]{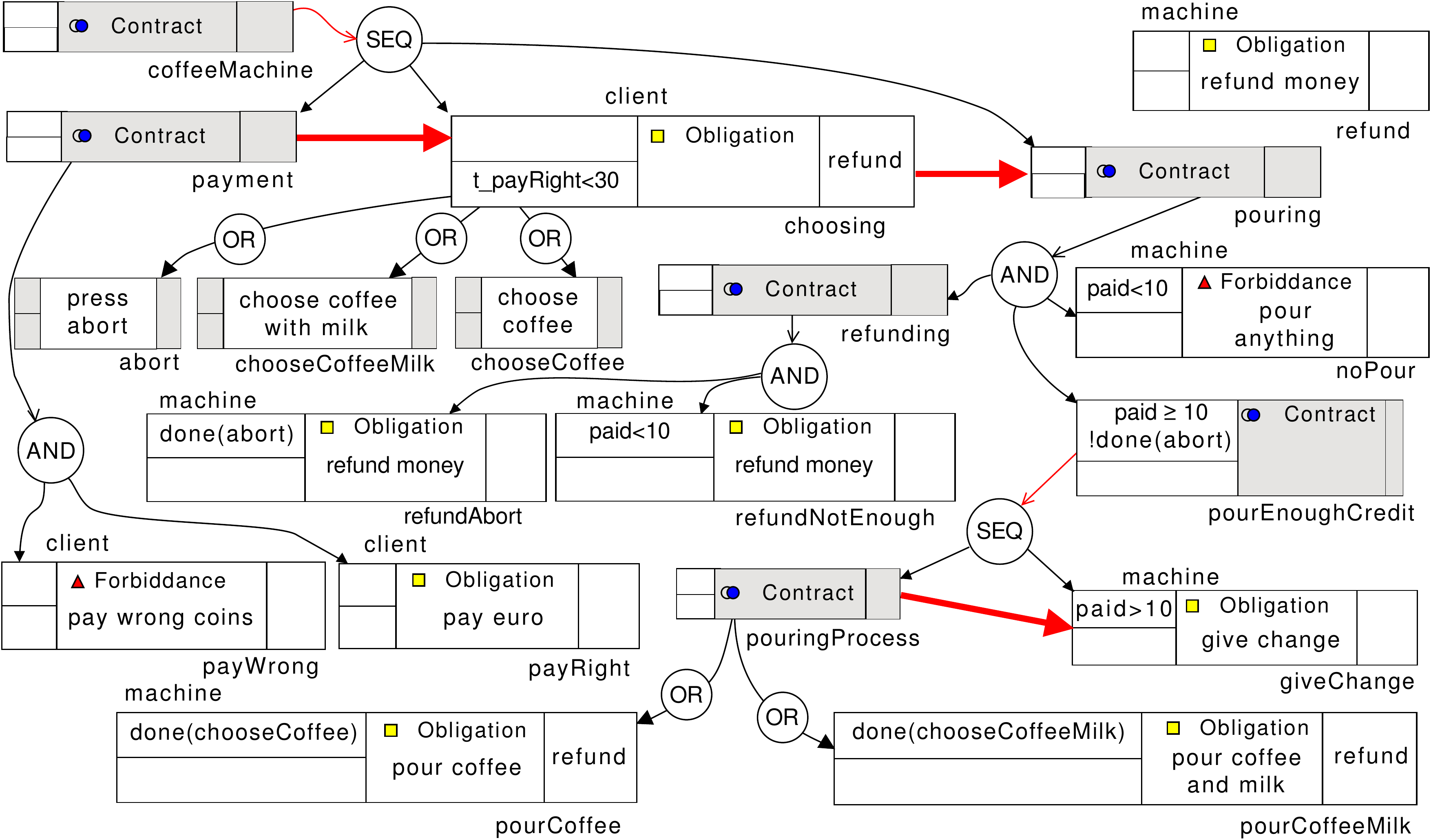}
  \vspace{0mm}
  \caption{The complete \codiag\ for the coffee machine example.}
  \label{fig:final-codiag}
\end{figure}

The \codiag\ for the final contract is shown in \autoref{fig:final-codiag}.
It includes the handling of the \conname{abort} action and gives an ordering to the sub-contracts.
Note how there are two separate contracts in the CNL verbalisation: \conname{coffeeMachine} and \conname{refund}, the latter being referenced as a reparation of the former.

The \codiag\ editor allows changes to be made locally while retaining the contract's overall structure, for instance inserting an additional option for a new beverage.
The CNL editor is instead most practical for replicating patterns or creating large structures such as sequences of clauses, that are faster to outline in text and rather tedious to arrange in a visual language.
The two editors have the same expressive power and the user can switch between them as they please.

\begin{figure}[p]
\begin{lstlisting}[numbers=left]
coffeeMachine : the following, in order
  - payment : payWrong : client mustn't pay wrong coins otherwise see refund and payRight : client is required to pay euro
  - choosing : when clock t_payRight less than 30 client is required
    - abort : to press abort , or
    - chooseCoffeeMilk : to choose coffee with milk , or
    - chooseCoffee : to choose coffee  otherwise see refund
  - pouring : all of
    - pourEnoughCredit : when abort is not done and variable paid not less than 10 first pouringProcess : pourCoffee : if chooseCoffee is done machine is required to pour coffee otherwise see refund or pourCoffeeMilk : if chooseCoffeeMilk is done machine is required to pour coffee and milk otherwise see refund , then giveChange : if variable paid greater than 10 machine is required to give change
    - noPour : if variable paid less than 10 machine mustn't pour anything
    - refunding : refundNotEnough : if variable paid less than 10 machine is required to refund money and refundAbort : if abort is done machine is required to refund money
refund : machine is required to refund money
\end{lstlisting}
\caption{The final verbalisation for the coffee machine example.}
\label{fig:final-cnl}
\end{figure}

\section{Evaluation}
\label{sec:evaluation}

\subsection{Metrics}

The GF abstract syntax for basic \codiags{} contains 48 rules, although the inclusion of large parts of the RGL for phrase formation pushes this number up to 251.
Including the large-scale English dictionary inflates the grammar to 65,174 rules.
As a comparison, a previous similar work on a CNL for the contract logic \CL\ \cite{ACS13fca} had a GF grammar of 27 rules, or 2,987 when including a small verb lexicon.

\subsection{Classification}

Kuhn suggests the PENS scheme for the classification of CNLs \cite{Kuhn2014}.
We would classify the CNL presented in the current work as P\textsuperscript{5}E\textsuperscript{1}N\textsuperscript{2-3}S\textsuperscript{4},\ F\ W\ D\ A.
P (precision) is high since we are implementing a formal grammar;
E (expressivity) is low since the CNL is restricted to the expressivity of the formalism;
N (naturalness) is low as the overall structure is dominated with clause labels and bullets;
S (simplicity) is high because the language can be concisely described as a GF grammar.
In terms of CNL properties, this is a written (W) language for formal representation (F), originating from academia (A) for use in a specific domain (D).

The P, E and S scores are in line with the problem of verbalising a formal system.
The low N score of between 2--3 is however the greatest concern with this CNL.
This is attributable to a sentence structure is not entirely natural,
somewhat idiosyncratic punctuation,
and a bulleted structure that could restrict readability.
While these features threaten the naturalness of the CNL in raw form,
we believe that sufficiently developed editing tools have a large part to play in dealing with the structural restrictions of this language.
Concretely, the ability to hide clause labels and fold away bulleted items can significantly make this CNL easier to read and work with.



\section{Related work}


\codiags\ may be seen as a generalisation of \CL\ \cite{Prisacariu2007fle,prisacariu9cl,PS12ddl} in terms of expressivity.%
\footnote{On the other hand, \CL\ has three different formal semantics:  an encoding into the $\mu$-calculus, a trace semantics, and a Kripke-semantics.}
In a previous work, Angelov et al. introduced a CNL for \CL\ in the framework \anacon\ \cite{ACS13fca}.
\anacon\ allows for the verification of conflicts (contradictory obligations, permissions and prohibitions) in normative texts using the CLAN tool \cite{fenech2009clan}.
The biggest difference between \anacon\ and the current work,
besides the underlying logical formalism,
is that we treat agents and actions as linguistic categories, and not as simple strings.
This enables better agreement in the CNL which lends itself to more natural verbalisations, as well as making it easier to translate the CNL into other natural languages.
We also introduce the special treatment of two-item co-ordination, and have a more general handling of lists as required by our more expressive target language.

Attempto Controlled English (ACE) \cite{fuchs99ace3Manual} is a controlled natural language for universal domain-independent use.
It comes with a parser to discourse representation structures and a first-order reasoner RACE \cite{Fuchs2012}.
The biggest distinction here is that our language is specifically tailored for the description of normative texts, whereas ACE is generic.
ACE also attempts to perform full sentence analysis, which is not necessary in our case since we are strictly limited to the semantic expressivity of the \codiag\ formalism.

Our CNL editor tool currently only has a basic user interface (UI).
As already noted however, it is clear that UI plays a huge role in the effectiveness of a CNL.
While our initial prototypes have only limited features in this regard, we point to the ACE Editor, AceRules and AceWiki tools described in \cite{Kuhn2010}
as excellent examples of how UI design can help towards solving the problems of writability with CNLs.


\section{Conclusion}

This work describes the first version of a CNL for the \codiag\ formalism, together with web-based tools for building models of real-world contracts.

The spreadsheet format mentioned in \autoref{fig:architecture} was not covered in this paper, but we aim to make it another entry point into our system.
This format shows the mapping between original text and formal model by splitting the relevant information about modality, agent, object and constraints into separate columns.
As an initial step, the input text can be separated into one sentence per row, and for each row the remaining cells can be semi-automatically filled-in using machine learning techniques.
This will help the first part of the modelling process by generating a skeleton contract which the user can begin with.


We plan to extend the CNL and \codiag\ editors with better user interfaces for easing the task of learning to use the respective representations and helping with the debugging of model errors.
We expect to have more integration between the two applications, in particular the ability to focus on smaller subsections of a contract and see both views in parallel.
While the CNL editor already has basic input completion,
it must be improvemed such that completion of functional keywords and content words are handled separately.
Syntax highlighting for indicating the different constituents in a clause will also be implemented.

We currently use the RGL \textit{as is} for parsing agents and actions without writing any specific constructors for them, which creates the potential for ambiguity.
While this does not effect the conversion process, ambiguity is still an undesirable feature to have in a CNL.
Future versions of the grammar will contain a more precise selection of functions for phrase construction, in order to minimise ambiguity.


Finally, it is already clear from the shallow evaluation in section \ref{sec:evaluation} that the CNL presented here suffers from some unnaturalness.
This can to some extent be improved by simple techniques, such as adding variants for keywords and phrase construction.
Other features of the \codiag\ formalism however are harder to linearise naturally, in particular mandatory clause labels and arbitrarily nested lists of constraints and actions.
We see this CNL as only the first step in a larger framework for working with electronic contracts,
which must eventually be more rigorously evaluated through a controlled usability study.


\section*{Acknowledgements}
The authors wish to thank the Swedish Research Council for financial support under grant nr. 2012-5746. 
We are also very grateful to the anonymous reviewers for their suggestions, in particular with regards to CNL evaluation and classification using the PENS scheme.

\bibliography{cnl2014}
\bibliographystyle{splncs}

\end{document}